\documentclass[10pt,twocolumn,letterpaper]{article}

\usepackage{cvpr}
\usepackage{times}
\usepackage{epsfig}
\usepackage{graphicx}
\usepackage{amsmath}
\usepackage{amssymb}

\usepackage{tabu}
\usepackage{url}

\cvprfinalcopy 

\graphicspath{}

\ifcvprfinal\fi
\begin{document}

\title{FotonNet: A HW-Efficient Object Detection System Using 3D-Depth Segmentation and 2D-DNN Classifier}

\author{Gurjeet Singh\\
{\tt\small singhg@oregonstate.edu}
\and
Sunmiao\\
{\tt\small 18112020006@fudan.edu.cn}
\and
Shi Shi\\
{\tt\small sshi15@fudan.edu.cn }
\and
Patrick Chiang\\
{\tt\small pchiang@oregonstate.edu}\\
}

\maketitle

\begin{abstract}
   Object detection and classification is one of the most important computer vision problems. Ever since the introduction of deep learning \cite{krizhevsky2012imagenet}, we have witnessed a dramatic increase in the accuracy of this object detection problem. However, most of these improvements have occurred using conventional 2D image processing. Recently, low-cost 3D-image sensors, such as the Microsoft Kinect (Time-of-Flight) or the Apple FaceID (Structured-Light), can provide 3D-depth or point cloud data that can be added to a convolutional neural network,  acting as an extra set of dimensions. In our proposed approach, we introduce a new 2D + 3D system that takes the 3D-data to determine the object region followed by any conventional 2D-DNN, such as AlexNet.  In this method, our approach can easily dissociate the information collection from the Point Cloud and 2D-Image data and combine both operations later.  Hence, our system can use any existing trained 2D network on a large image dataset, and does not require a large 3D-depth dataset for new training.   Experimental object detection results across 30 images show an accuracy of 0.67, versus 0.54 and 0.51 for RCNN and YOLO, respectively.
   
\end{abstract}

\section{Introduction and Motivation}

The ability for robots and computers to see and understand the environment is becoming a burgeoning field, needed in autonomous vehicles, augmented reality, drones, facial recognition, and robotic helpers. Such systems require the detection and classification of objects to perform various tasks. In 2012,Krizhevsky et. al \cite{krizhevsky2012imagenet} \cite{denglarge} introduced the CNN (Convolutional Neural Network) based Deep Neural Network technique for image classification which outperformed existing techniques. After the successful implementation of image classification, Ren et. al. \cite{ren2015faster} and implemented Region-Based CNN for Object Detection. The Fully Convolutional Network provided a solution to semantic segmentation, followed by Mask RCNN \cite{long2015fully}, which further improved the capabilities of the Deep Neural Network.\\

All of the previously aforementioned networks were designed assuming a standard 2D-image dataset, due to the wide prevalence of low-cost 2D-image sensors (i.e.  cellphones with open-source sharing of their RGB-data images).  Recently, low-cost 3D-image sensors that can be embedded into cellphones (i.e. FaceID, Google Tango) have begun arising, enabling the potential for ubiquitous RGB+D sensor data, that can be used to improve object detection.  Previous work from Gupta et. al. \cite{gupta2014learning} trained a Faster-RCNN on RGBD dataset NYUD\cite{silberman2012indoor}, using HHA encoding of 3D-depth images with 2D-images. This approach consider depth data as an additional NN input and not as independent values coming from different sensors, and thereby needs a large amount of depth data to train the neural network. Unfortunately, since 3D-depth sensors are not widely available as of yet, the ability to create a clean, accurate, and well-annotated RGB + Point Cloud dataset ($\sim$ 5000 images) as large as ImageNet ($\sim$ 1.2 million images) is still not possible. Unavailability of depth sensors also limits researchers from testing a large amount of corner cases (i.e. difficult scenarios such as different SNR situations for 2D-images or 3D-depth data.)\\  Hence, the accuracy of 
a CNN trained with a small amount of RGB+D data will be inferior to a CNN trained with a large amount of RGB data.


\section{Previous Work}
In this section, we overview and discuss some previous object detection techniques and network architectures used for images and depth datasets. \\
\subsection{Faster R-CNN}
Ren et. al.\cite{ren2015faster} proposed a Region Proposal Network (RPN) to obtain bounding region boxes. These boxes are then sent to a classification layer to decide whether it contains  any object or not. Since there are overlapping regions of bounding boxes, non max suppression (NMS) is used to combine different adjacent bounding boxes.\\

\begin{figure}[h]
\includegraphics[width=8cm]{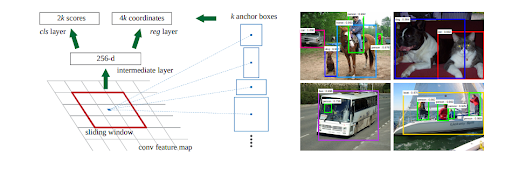}
\caption{Region Proposal Network and the Output from Pascal VOC }
\label{fig:FRCNN}
\end{figure}

For their experiment, both RPN and classification networks share the VGG-16 model (13 CNN and 3 FCN layers) for feature extraction and the NMS limit to 0.7, resulting in bounding boxes of 2000 for each image. Gupta et. al. \cite{gupta2014learning} used Faster RCNN to learn rich features from RGBD images of NYUD dataset.They transformed depth information into 3-channels into HHA transformation.  Figure \ref{fig:FRCNN} shows the basic operation of Faster R-CNN, showing the anchor boxes using the Region Proposal Network(RPN), and then the final output.\\

\subsection{Mask R-CNN}
He et. al.\cite{he2017mask} demonstrates Mask R-CNN which is an extension of Faster R-CNN. Mask R-CNN extends Faster R-CNN by adding an object mask detector in parallel, which is used for instance segmentation.\\

\begin{figure}[h]
\includegraphics[width=8cm]{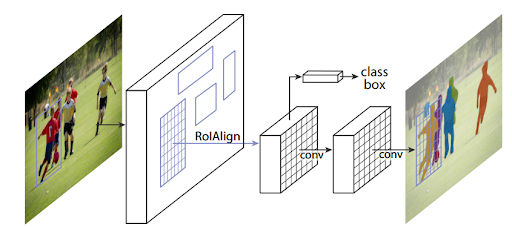}
\caption{Mask R-CNN architecture for instance segmentation }
\label{fig:MaskRCNN}
\end{figure}

Faster RCNN does not provide pixel representation of object in a scene. Instead, it just provides a bounding around an object. To address this problem, He et al. uses a ROI align layer that gives pixel to pixel representation from the bounding boxes. Furthermore, there is a ROI pool layer that combines local bounding boxes. Figure \ref{fig:MaskRCNN} demonstrates the ROI pool layer and instance segmentation layers used for instance segmentation that operates on top of Object Detection. \\

\subsection{You Only Look Once (YOLO)}
\begin{figure}[h]
\includegraphics[width=8cm]{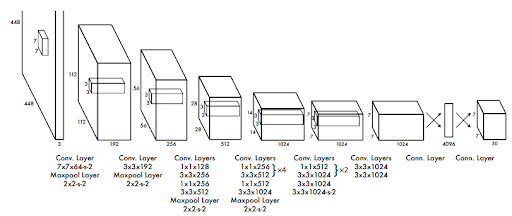}
\caption{YOLO architecture \cite{redmon2016you}}
\label{fig:YOLO}
\end{figure}

Redmon et. al. \cite{redmon2016you} proposes a much simpler neural network for Object Detection. This network does not need Region Proposal Network (RPN) layer for training. Instead, the picture is divided into several predesigned anchor boxes. Next, Redmon et. al. run classifiers such that overlapping bounding boxes are removed using Non-max suppression (NMS). The advantage of this network lies in its end-to end training. In this way, Redmon et. al. can train networks faster and easily by sacrificing some accuracy. Figure \ref{fig:YOLO} shows a detailed YOLO architecture with 24 convolutional layers and 3 fully connected layers.\\


\subsection{VoxelNet}
\begin{figure}[h]
\includegraphics[width=8cm]{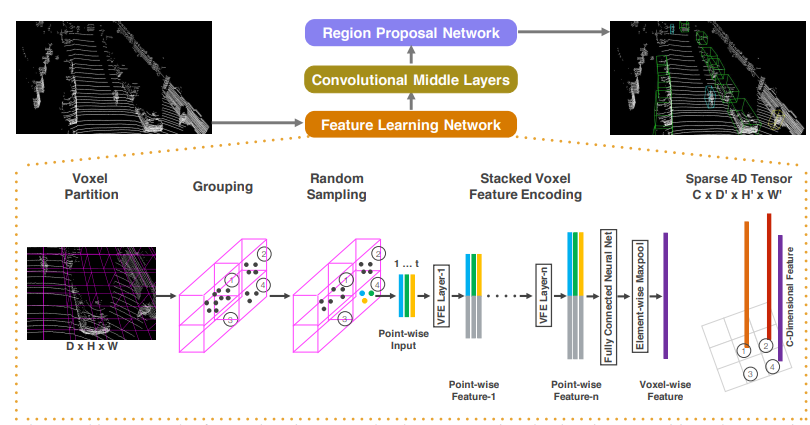}
\caption{VoxelNet architecture \cite{zhou2017voxelnet}}
\label{fig:VoxelNet}
\end{figure}

VoxelNet \cite{zhou2017voxelnet} is a unique way of running object detection just on 3D point-cloud data (with no other sensor data input). Point Clouds are sparse, such that points clouds are converted to voxels and therefore just samples with certain chosen thresholds. Next, random samples are chosen and converted to point-wise inputs for feature learning.  Figure \ref{fig:VoxelNet} shows various layers and final output for VoxelNet architecture.\\

\section{Shortcomings of Prior Art}
Most of the networks previously described use only conventional 2D image datasets. NYUD processes RGB+D data, but unfortunately the dataset exhibits a very high density of cluttered objects, with many of these objects exhibiting no depth features (curtains, windows, etc.) Unfortunately, with such a cluttered RGB+D dataset, any training using RGB data and depth data will not contribute much to learning. For example, we previously tried to use the depth data of NYUD for training, but because there are a little amount of features from depth, it is very difficult to stop overfitting because of the small number of images. \\

In regards to VoxelNet, the network is trained with a number of different objects (Car, Bike,etc), such that training for another network with a different set of objects is impractical.  For both NYUD and VolxelNet, a large amount of 3D depth data for reliable training of neural network is required. Unfortunately, there currently exists no 3D-depth dataset large enough to rival conventional 2D-image datasets (ImageNet etc.)\\

\section{Proposed Work}
In this paper, we propose a DNN Object Detection system that dissociates the depth-data from the RGB data. In this way, our proposed system does not need require a large training dataset for the depth data but still simultaneously extracts meaningful information from the depth sensor. Furthermore, our system combines this depth data with conventional 2D-trained image data to generate a practical, low-complexity object detection and classification system. \\ 

This paper is constructed as follows. First we will give a system overview of our system, followed by test results, a comparison with current state-of-the-art systems, and then concluding with the our future research direction.\\

\begin{table*}[t]
\begin{center}
\caption{Summary of Proposed Work and Comparison with Previous Work}
\label{table:1}
\begin{tabular} {|m{2cm}|m{2cm}|m{2cm}|m{2cm}|m{2cm}|m{2cm}|} 
 \hline
 \textbf{Neural Network} & \textbf{Training Dataset} & \textbf{Accuracy} & \textbf{Latency} & \textbf{Complexity} & \textbf{HW Implementation Complexity} \\ 
 \hline\hline
 Faster R-CNN/Mask RCNN & PASCAL-VOC, MSCOCO & High & High & High & High \\ 
 \hline
 YOLO  & PASCAL-VOC, MSCOCO & Average & Low & Low & High \\
 \hline
 VoxelNet  & KITTI & Average & --- & High & High \\
 \hline
 \textbf{FotonNet(this work)} & \textbf{Pretrained- AlexNet} & \textbf{High} & \textbf{Low} & \textbf{Low} & \textbf{Low} \\ 
 \hline
\end{tabular}
\end{center}
\end{table*}

\section{System Overview}
\begin{figure}[h]
\includegraphics[width=8cm]{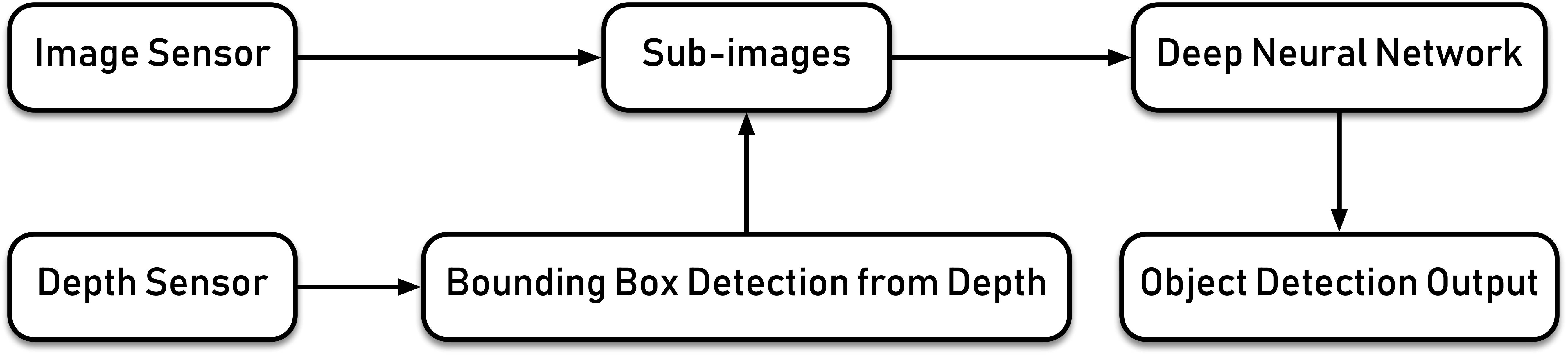}
\caption{FotonNet System Architecture}
\label{fig: SystemArchitecture}
\end{figure}

Figure \ref{fig: SystemArchitecture} shows an overview of the system architecture.  Our system was designed with the understanding that there currently exists no large dataset of depth data for training. First, our system generates bounding boxes from the depth sensor output by performing clustering on the 3D point cloud.  After denoising and clean up of the clustered depth objects, we use the clustered objects to split the 2D image into sub-images, which are finally fed  into a 2D-Image Deep Neural Classification Network. A detailed operation for each individual subsystem is explained below.\\ 

\subsection{Bounding Box Detection}
\begin{figure}[h]
\includegraphics[width=8cm]{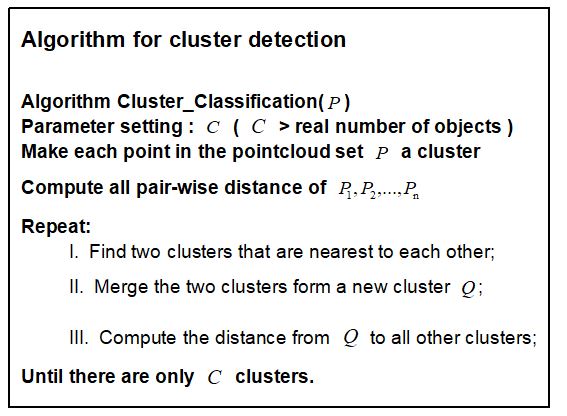}
\caption{Algorithm for Clustering}
\label{fig: ClusteringAlgorithm}
\end{figure}

The bounding box detector takes the 3D point cloud input from the depth sensor and generates bounding boxes, based upon the different clusters of point cloud data. In our system, we use  the Hierarchical Clustering Algorithm, shown in Figure \ref{fig: ClusteringAlgorithm}, due to its ease of implementation. Hierarchical clustering is a method of cluster analysis commonly used in data mining and statistics that builds upon a hierarchy of clusters \cite{ward1963hierarchical}. \\

\begin{figure}[h]
\includegraphics[width=8cm]{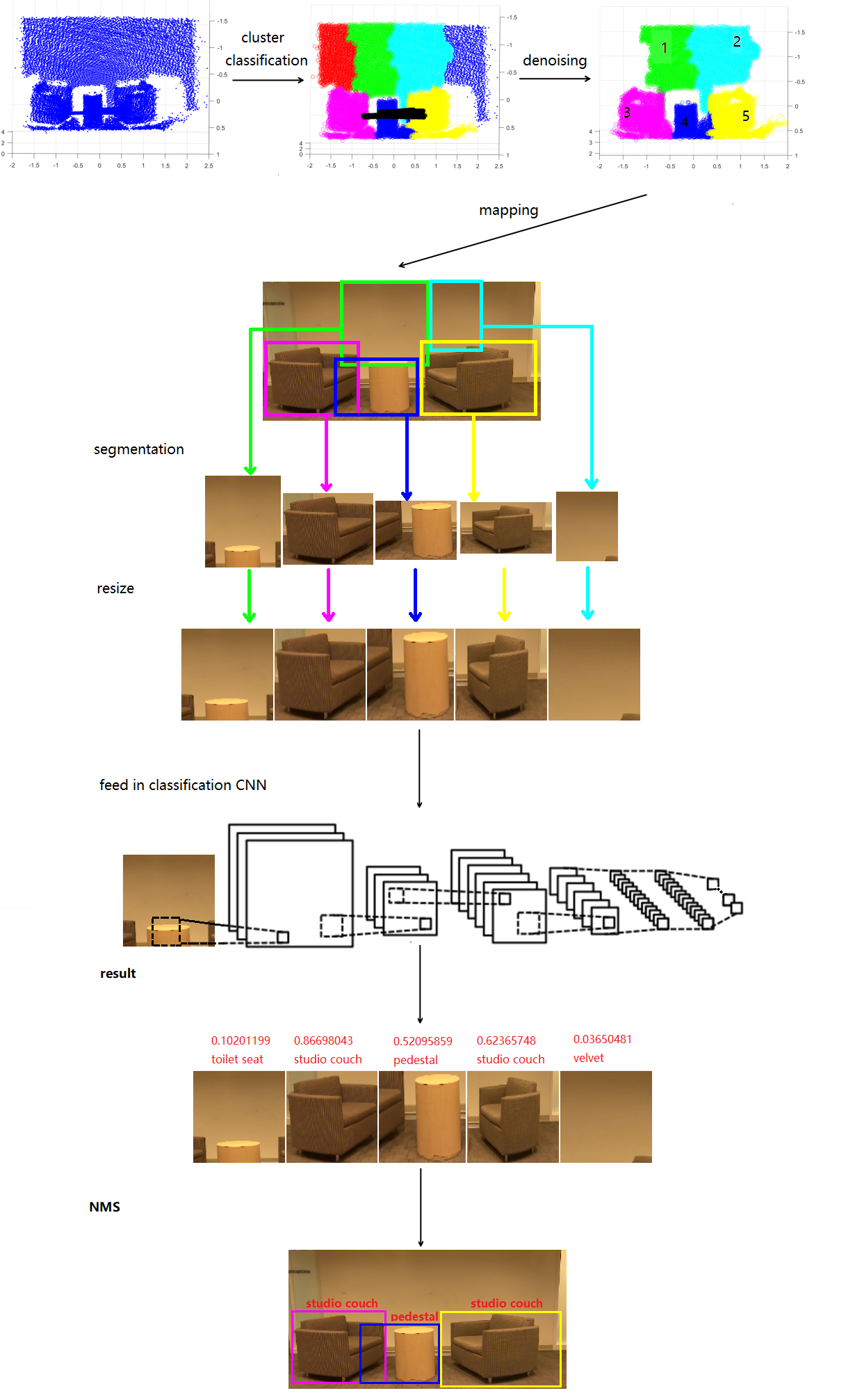}
\caption{Architecture for FotonNet}
\label{fig: GeneralArchitecture}
\end{figure}

\subsection{Denoising}
After the clustering step, several simple denoising steps are implemented that act as a pre-processing and image cleanup that improves the following object detection.  First, the two clusters from the two top corners are eliminated since they are likely peripheral background images and not the main images for classification. Second, small-sized clusters which are a small percentage ratio of the entire point-cloud are judged as noise (not real objects), and are also subtracted. Several other subimages that are extraneous, such as point-clouds from the wall and the ground, have not been eliminated in this preliminary implementation.\\

\subsection{Sub-Images}
After the detection of several different bounding boxes, we split the image into several sub-images so that we can run object detection on several objects detected by the depth sensor. The 3D-depth sensor is calibrated with the 2D-camera such that the bounding boxes can be projected on to the 2D-image plane, using the method from Park et. al. \cite{park2014calibration}. In particular, we utilize Park's polygonal method  and calculate the Projection Matrix for Singular Value Decomposition (SVD).\\

\subsection{Deep Neural Network}
Our proposed Object Detection system has the advantage in that it can use any existing classification network that has been trained on large datasets of 2D images (as opposed to VoxelNet or NYUD). The advantage of this system is that we can use an already pretrained network and therefore do not need to train a new network for depth data. For this initial experiment, we utilize AlexNet due to its widespread availability.\\

\begin{figure*}[t]
\includegraphics[width=18cm]{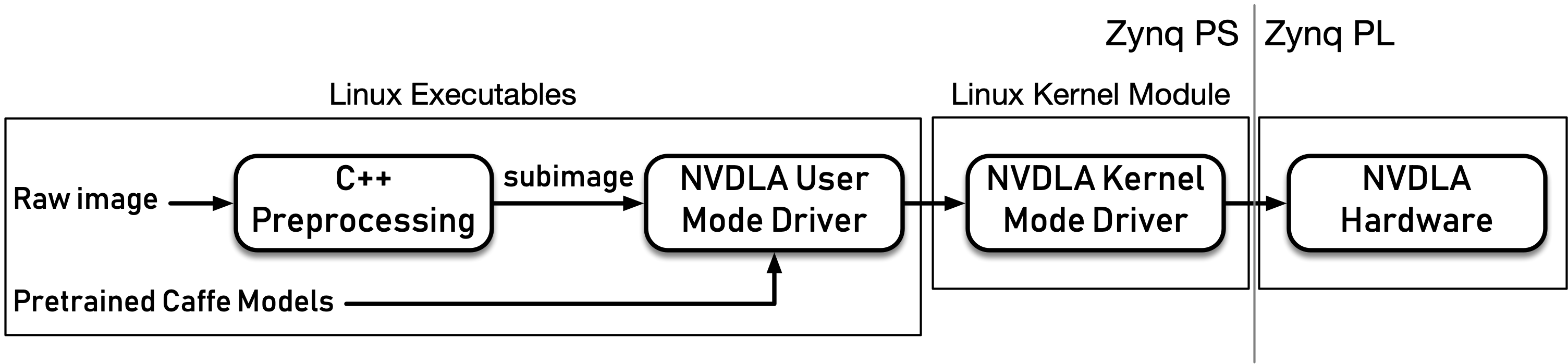}
\caption{Implementation Architecture}
\label{fig: ImplementationWorkflow}
\end{figure*}

One advantage for our system is that it implements  AlexNet inference using the NVDLA open-source deep-learning inference engine \cite{NVDLAPrimer}, thereby enabling a future low-cost ASIC implementation. For our system verification, we used a ZYNQ ultrascale+ SoC to run the entire system, which includes the image clustering, preprocessing, and AlexNet inference. The preprocessing part is written in C++, compiled with the gcc-linario arm-linux cross compiler. Subimages are generated accordingly and fed into the NVDLA User Mode Driver.  The NVDLA small configuration hardware is programmed into the programmable logic (PL) part of the Zynq SoC. A NVDLA full configuration can run AlexNet at 1100 Frames per second with 2048 Multiply-accumulate units (MACs) when its batch size is 16 \cite{NVDLAOpensourcePerformance}. The NVDLA small configuration we are using exhibits 64 MACs, which is 32 times less hardware than the full configuration, as it is not the bottleneck of speed for our implementation. The implementation workflow of the entire system is shown in Figure \ref{fig: ImplementationWorkflow}. \\

\subsection{Object Detection Output}
After Classification, we obtain the output with corresponding bounding boxes. We apply a threshold \(\lambda\)  on the final subimages in order to determine if there are any extra bounding boxes. The classification probability below a certain threshold (i.e. 0.2) will be discarded.\\

\section{Experiment}
\begin{figure}[h]
\includegraphics[width=9cm]{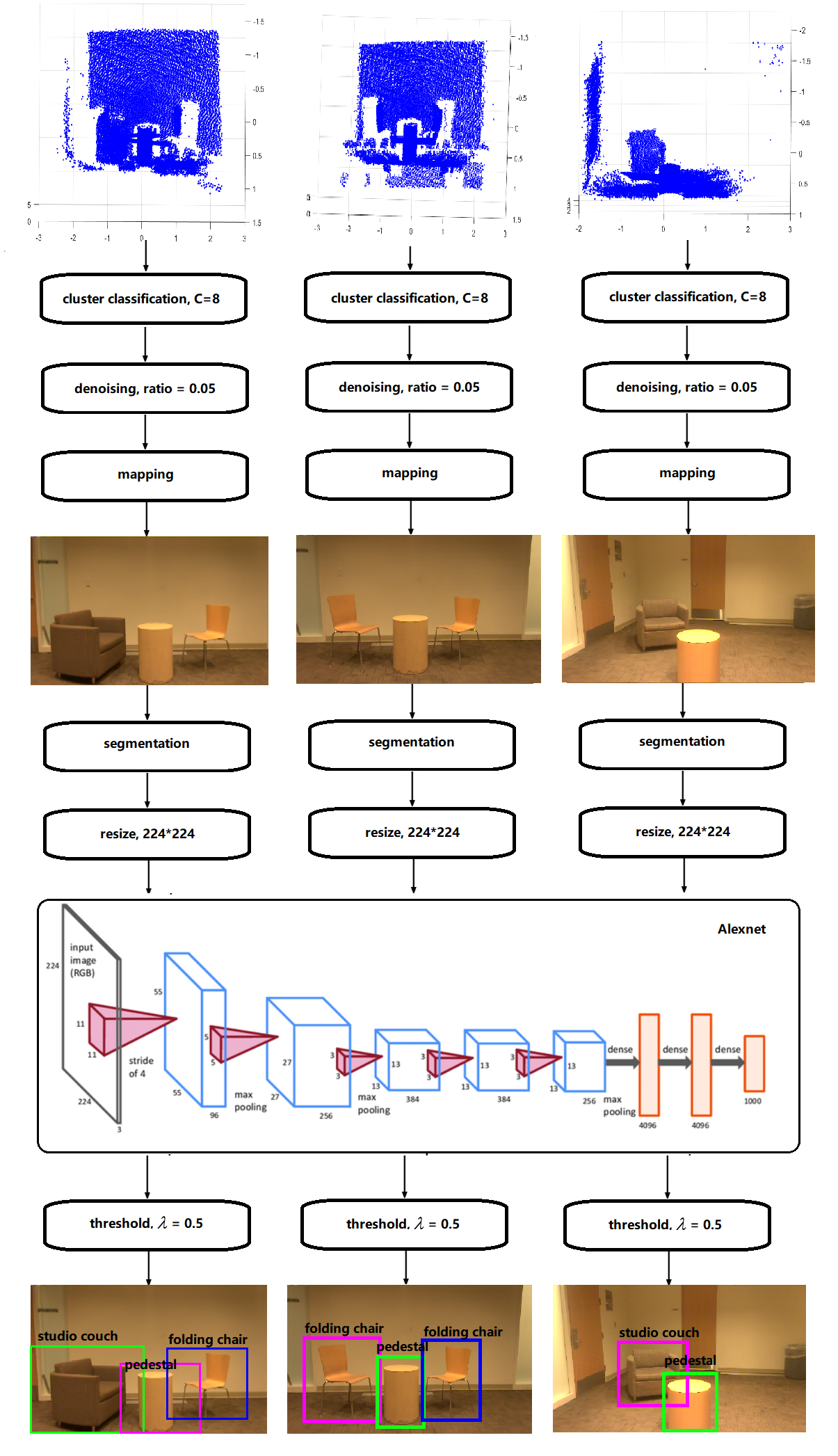}
\caption{Final output of the system}
\label{fig: FinalOutput}
\end{figure}

Since the depth data is used here to perform clustering, it is difficult to compare our system  with previous benchmarks. In order to perform a side by side comparison of our system versus other algorithms, we captured both the 2D-Image and 3D-Depth data of 30 images with various indoor objects inside. We tested our proposed FotonNet using these 30 images and point cloud data, while for the conventional networks \cite{Detectron2018} and YOLO darknet \cite{yolov3}, we only used the RGB images from the scenes and ran through detection. The output of FotonNet is shown in Figure \ref{fig: FinalOutput}.\\

\section{Results}
We tested this system with our 30 test images and calculated the mean Average Precision(mAP). The mAP for our system is 0.72413, with a measured desktop latency of  5.99655s. The results of our dataset and the parameters sets are as follow:
\begin{itemize}
  \item Data Processing CPU Time = 5.96655s (python)
  \item AlexNet Forward:  30.1214 ms
  \item AlexNet Backward: 41.7727 ms
  \item Total: 72.5577 ms (GPU)
\end{itemize}

\section{Comparison Table}
In order to compare our system versus others, we ran our test images through each of the networks, thereby providing the IOU, latency and size of each network. We also provide the latency and accuracy as reported by existing neural networks.

For the AlexNet network using the Caffe2 framework with a Nvidia 1080 Ti GPU, the forward propagation time is 4msec. Speed of our system then will depend on then number of clusters we generate.  Assuming on average we generate 8 clusters, the speed of our system is approximately 30fps. Note that it is assumed that the clustering algorithm is available and integrated into the on-board 3D-depth sensor. However, the speed can be easily increased if we use the NVDLA classification engine, which reports a speed of 1100 fps.

One major advantage of FotonNet is its ease-of-deployment. By dissociating depth from the RGB data, it will be easier to expand object detection to even more classes and neural networks then just limited Object Detection classes. This makes it easier to implement and expand than currently existing systems.

\begin{figure*}[t]
\includegraphics{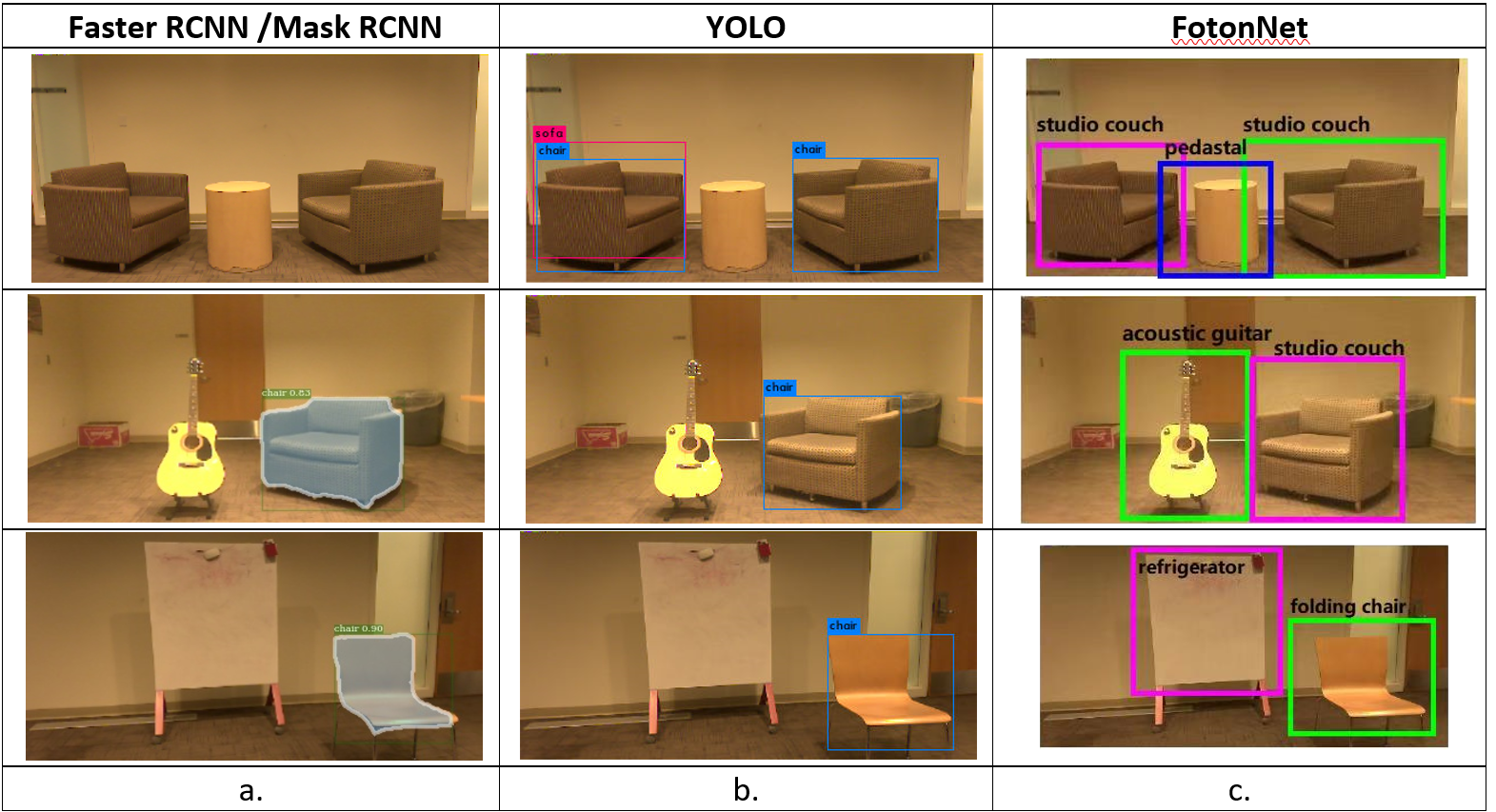}
\caption{a. Sample output from Faster RCNN /Mask RCNN b. Sample output from YOLOv3 c. Sample output from FotonNet}
\label{fig: Comparison}
\end{figure*}

\begin{table*}[t]
\begin{center}
\caption{Comparison of System with Existing State of the art methods}
\label{table:2}
\begin{tabular} {|m{2cm}|m{2cm}|m{2cm}|m{4.5cm}|m{2cm}|m{2cm}|} 
 \hline
 \textbf{Network} & \textbf{Usability} & \textbf{Rough Estimate of HW-Complexity} & \textbf{Accuracy} & \textbf{Latency (FPS)} & \textbf{No. of training Parameters} \\ 
 \hline\hline
 Faster RCNN Mask RCNN & No training data for depth & Large Network & 0.759 (COCO test (mAP)) \newline 0.53968 (30 images(IOU)) & 7 & VGG16 + RPN + Regression Layer \\ 
 \hline
 YOLO  & No training data for depth & Large Network & 0.634 (COCO test(mAP) \newline 0.506849 (30 images(IOU))& 21 & 24 Conv Layer + 2 FC Layer \\
 \hline
 FotonNet  & Need need to training when using large classification network  & Low Complexity, Cluster and Classification can be moved on chip & ---- (COCO Test) \newline 0.67 (30 images(IOU))
 & ~30(Assuming Clustering is performed by sensor locally) & Using Pre- Trained AlexNet \\
 \hline
\end{tabular}
\end{center}
\end{table*}

\section{Conclusion}
We demonstrate a new HW-system architecture for object detection which leverages low-cost 3D depth sensors. The size of this network is significantly reduced compared with the status quo because of the reduction in the number of ROIs and the elimination of any extra training steps needed for object classification. We used this approach to keep in mind that we are going to move Boundary box detection on-chip in Depth Sensor. We found that our approach outperform YOLO and Faster-RCNN in object detection. 

In this paper, we experimented with hierarchical clustering algorithm. We will try to come up with a faster clustering algorithm and try to migrate that functionality on chip.

{\small
\bibliographystyle{ieee}
\bibliography{egbib}
}

\end{document}